*Research article*

# Improved YOLOv7 model for insulator defect detection

Zhenyue Wang[1,2], Guowu Yuan[1,*], Hao Zhou[1], Yi Ma[2], Yutang Ma[2] and Dong Chen[1]

[1] School of Information Science and Engineering, Yunnan University, Kunming 650504, China
[2] Electric Power Research Institute, Yunnan Power Grid Co., Ltd, Kunming, Yunnan 650214, China

**\* Correspondence:** Email: gwyuan@ynu.edu.cn.

**Abstract:** Insulators are crucial insulation components and structural supports in power grids, playing a vital role in the transmission lines. Due to temperature fluctuations, internal stress, or damage from hail, insulators are prone to injury. Automatic detection of damaged insulators faces challenges such as diverse types, small defect targets, and complex backgrounds and shapes. Most research for detecting insulator defects has focused on a single defect type or a specific material. However, the insulators in the grid's transmission lines have different colors and materials. Various insulator defects coexist, and the existing methods have difficulty meeting the practical application requirements. Current methods suffer from low detection accuracy and mAP0.5 cannot meet application requirements. This paper proposes an improved you only look once version 7 (YOLOv7) model for multi-type insulator defect detection. First, our model replaces the spatial pyramid pooling cross stage partial network (SPPCSPC) module with the receptive filed block (RFB) module to enhance the network's feature extraction capability. Second, a coordinate attention (CA) mechanism is introduced into the head part to enhance the network's feature representation ability and to improve detection accuracy. Third, a wise intersection over union (WIoU) loss function is employed to address the low-quality samples hindering model generalization during training, thereby improving the model's overall performance. The experimental results indicate that the proposed model exhibits enhancements across various performance metrics. Specifically, there is a 1.6% advancement in mAP_0.5, a corresponding 1.6% enhancement in mAP_0.5:0.95, a 1.3% elevation in precision, and a 1% increase in recall. Moreover, the model achieves parameter reduction by 3.2 million, leading to a decrease of 2.5 GFLOPS in computational cost. Notably, there is also an improvement of 2.81 milliseconds in single-image detection speed. This improved model can detect insulator defects for diverse materials, color insulators, and partial damage shapes in complex backgrounds.

---

## 1. Introduction

Insulators are essential components of power transmission lines, playing a crucial role in insulation and structural support and ensuring the safety of transmission lines. In complex terrain and harsh environments, most transmission lines are subjected to physical and chemical erosion over time, making them prone to defects such as self-explosion and damage. Regular inspections are necessary to monitor the operational status of insulators and ensure the safe operation of power transmission lines [1]. Insulator defects in power line inspections require manual observation or manual detection of drone images. This approach is labor-intensive, inefficient, and easily influenced by the skill level of the personnel, resulting in inconsistent detection results [2]. Relying solely on manual inspection is insufficient to meet the requirements of automated grid inspections [3]. Nowadays, drones can capture image data of power transmission lines, and computer vision methods can identify images automatically. This approach effectively solves manual inspections and is one of the crucial methods for line inspections [4,5].

Early detection of insulator defects was primarily based on traditional digital image processing techniques, using traditional template matching methods to extract defect features from insulator images [6,7], and employing manually designed templates to extract insulator features [8]. Nevertheless, the intricate and varied intricacies present in insulator images pose challenges in encapsulating the holistic attributes of the targets through a singular manually crafted descriptor. Consequently, defect detection becomes arduous. Additionally, employing machine learning techniques grounded in probabilistic statistics frequently demands intricate feature descriptors. However, feature representations acquired from shallow network architectures exhibit constrained efficacy and generalization capacity in tackling intricate object detection issues.

With the advancement of deep learning, several approaches for detecting insulator defects have emerged. Jiang et al. [10] utilized a multi-layer perceptron ensemble learning model, achieving a detection accuracy of 92.26%. Sadykova et al. [11] applied a YOLOv2 model for insulator detection, but detecting defects still required manual inspection. Wang et al. [12] proposed an enhanced faster region-based convolutional neural network (R-CNN) model, integrating split-attention networks (ResNeSt) and introducing the RPN network to enhance defect feature extraction, resulting in an accuracy of 98.38%. To address limited training data and labor-intensive annotation, Shi et al. [13] introduced a weakly supervised learning method based on faster R-CNN, reaching a defect detection accuracy of 92.86% and an F1-Score of 90.85%. Zhao et al. [14] employed a feature pyramid network (FPN) FRCN for detecting dropout and crack defects on insulators. Luo et al. [15] proposed a combined object detection framework comprising Faster R-CNN and YOLOv3, achieving a reduced miss rate in insulator defect detection. Zhang et al. [16] designed a densely connected FPN, enhancing semantic and positional information integration, thus improving detection performance. Liu et al. [17] utilized YOLOv4 for insulator object detection, integrating the watershed algorithm for positioning insulator burst defects. Additionally, they proposed an improved YOLOv3 model, incorporating a densely connected convolutional networks (DenseNet) module, which achieved a detection accuracy of 96.29%. Wang et al. [19] enhanced the generative adversarial network (GAN) for insulator defect detection, resulting in an average precision of 84.6% for detecting glass insulator self-explosion defects. Kang et al. [20] introduced the concat bidirectional feature pyramid network (CAT-BiFPN) into YOLOv7 for multi-class insulator defects, improving feature fusion and achieving an average accuracy of 93.9%. Singh et al. [21] proposed a three-step method for classifying defective insulators in high-voltage transmission lines. Souza et al. [22] presented a hybrid version of YOLO called Hybrid-YOLO, combining YOLOv5 and ResNet-18 for object detection and classification, achieving an F1_score of 0.96216 and a mAP_0.5 of 0.99262. Stefenon et al. [23] introduced YOLOu-Quasi-ProtoPNet, a model trained from scratch, achieving an F1_score of 0.95165 based on DenseNet-161.

Currently, most research on insulator defect detection focuses on detecting a specific insulator defect or a particular material insulator. However, insulators in power grid transmission lines have different colors and materials, and multiple kinds of insulator defects coexist, and existing methods struggle to meet practical application requirements. The YOLOv7 [24] model is a relatively new object detection model that offers accuracy and speed advantages. Taking into account the intricacies of insulator defect detection, this study presents an enhanced multi-type insulator defect detection framework leveraging YOLOv7. Initially, the SPPCSPC module within the YOLOv7 architecture is substituted with the RFB module [25]. Subsequently, a coordinate attention mechanism [26] is integrated into the head section alongside the incorporation of a WIoU loss function [27]. The experimental results show that our refined model significantly improves the precision in identifying self-explosion defects and partial damage in power transmission line insulators.

The primary contributions of this study can be summarized as follows:

(1) We curate a dataset encompassing three distinct categories of insulator samples: self-explosion defects, partial damage, and normal insulators. Yunnan Power Grid Co., Ltd. of China Southern Power Grid, generously provides the images. These samples exhibit intricate backgrounds, diverse shooting perspectives, and various insulator characteristics, including shapes, materials, colors, and defect types.

(2) We introduce a novel multi-type insulator defect detection model built upon the enhanced YOLOv7 architecture, delivering superior accuracy in detection performance.

## 2. Dataset

Yunnan Power Grid Co., Ltd. provides the insulator defect dataset, and drones capture the sample images. We classify the insulator sample images into three categories: normal, self-explosion defects, and partial damage defects, with 1000 images for each type. Figure 1(a) shows normal insulators, Figure 1(b) shows self-explosion insulators, and Figure1(c) shows partially damaged insulators.

Because the sample images have large sizes with a resolution of 4000 × 3000 pixels and the detected insulator is only a tiny proportion of the whole image, it isn't easy to see the actual label after detection. In Figure 1, we only intercepted the defective part of the insulator and magnified it for clear display in the paper.

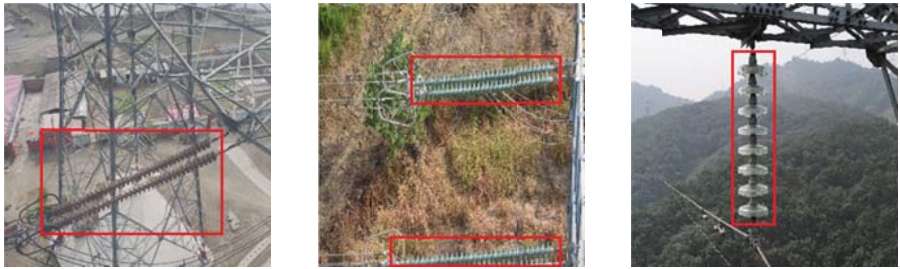

（a）Normal insulators

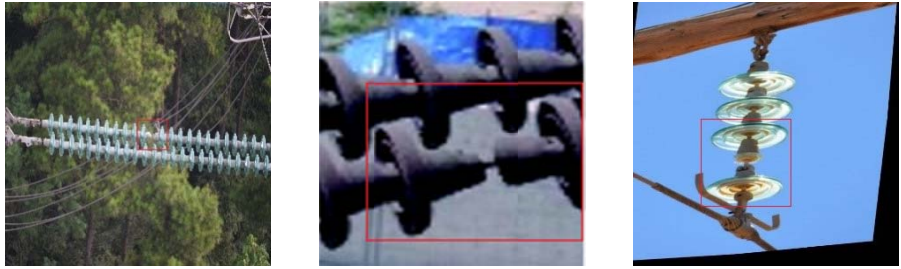

（b）Self-explosion defect insulators

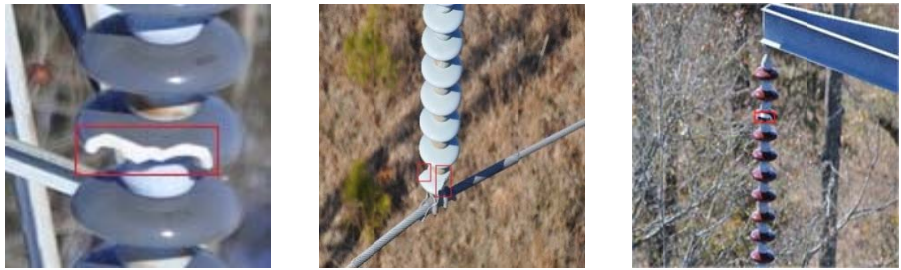

（c）Partial damage insulators

**Figure 1.** Collected insulator image dataset.

From Figure 1, the insulator sample images have complex background and the insulators exhibit various shapes, materials, and colors. Significant differences existed in shooting angles and sizes, and the same type of defect can have diverse manifestations. These diversities increase the difficulty of defect detection.

## 3. Proposed method

To tackle the complexities arising from the diverse materials, colors, intricate backgrounds, and varied damage shapes observed in insulators, we propose an enhanced model built upon the YOLOv7 framework. Within this model, the SPPCSPC module undergoes substitution with the RFB module, thereby augmenting the network's feature extraction capacity. Additionally, we integrate the coordinate attention (CA) mechanism into the head segment to enhance detection precision concerning small-scale self-explosion defects and partial damage instances, thus bolstering the network's feature representation capabilities. Furthermore, the WIoU loss function is introduced to mitigate the impact of low-quality samples on the model's training process, thereby enhancing its generalization performance. These enhancements are elaborated upon extensively in sections 3.2–3.4.

*3.1. Improved YOLOv7 model*

Figure 2 shows the improved YOLOv7 model, where the two red boxes indicate the modified parts in the original model. Specifically, Section 3.2 explains the receptive filed block (RFB) module, Section 3.3 describes the CA mechanism, and Section 3.4 presents the WIoU loss function.

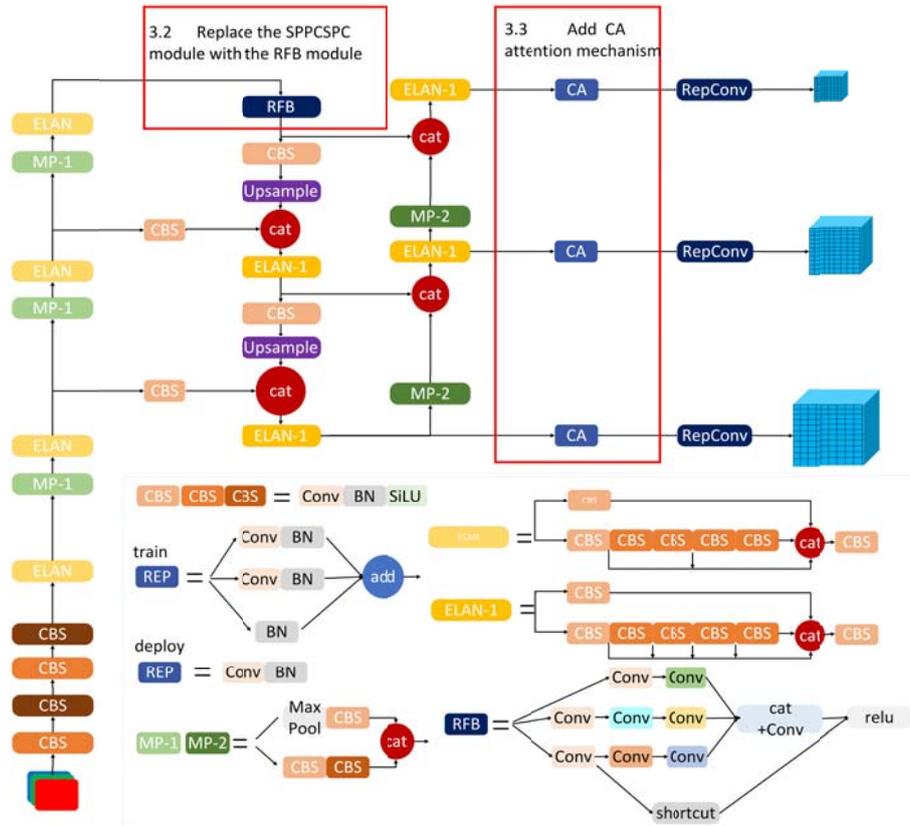

**Figure 2.** Improved YOLOv7 model for insulator defect detection.

## 3.2. Receptive field block (RFB)

Given the diverse materials and colors of insulators, along with the intricate backgrounds and shapes of partial damage, we opt to replace the SPPCSPC module found in the original YOLOv7 model with the RFB module to augment the network's feature extraction capabilities. While the SPPCSPC module was initially introduced within the YOLOv7 model, the RFB module takes a different approach. It employs multiple branch convolutional layers, utilizing convolutional kernels of various sizes to achieve superior performance. This approach aims to mimic the receptive fields of human vision, thereby enhancing the network's feature extraction prowess.

Inspired by the Inception concept [28], the RFB module distinguishes itself by incorporating a regular convolution followed by a dilated convolution [29] within each branch. Notably, the main convolutional kernel sizes and dilated factors vary across branches, contributing to a diverse range of feature extraction capabilities. The dilated convolution expands the receptive field while keeping the parameter count constant, facilitating the extraction of higher-resolution features. The RFB module is visually depicted in Figure 3.

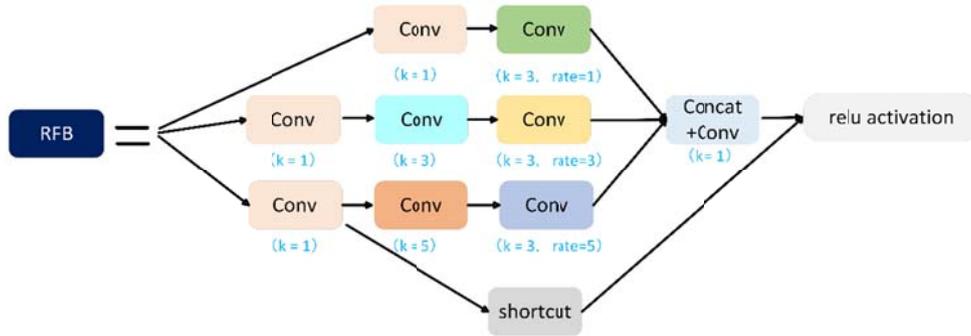

**Figure 3.** RFB structure.

*3.3.CA mechanism*

This study integrates a coordinate attention (CA) mechanism into the head section to mitigate the issue of missing small-scale self-blasts and damaged insulators. This mechanism effectively captures channel relationships and long-term dependencies by leveraging precise positional information. The CA mechanism operates through two primary steps: embedding coordinating information and generating coordinating attention. The CA attention block is illustrated in Figure 4.

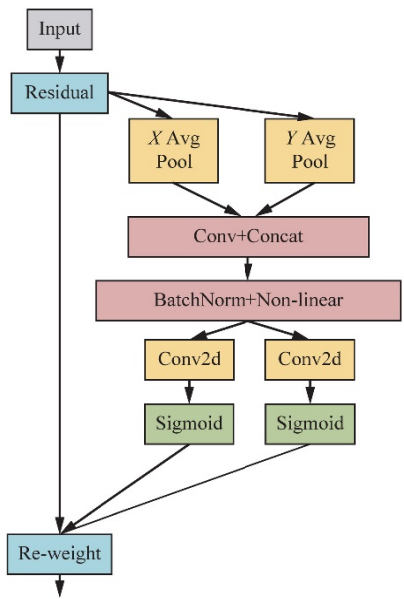

**Figure 4.** CA block.

The process of embedding coordinate information aims to tackle the challenge of condensing global spatial information into a single channel descriptor during global pooling, which can lead to the loss of positional details. To address this issue, the embedding step divides global pooling into horizontal and vertical direction encoding for each channel individually, as depicted in Eq (1). This transformation effectively converts the pooling operation into a one-to-one feature encoding process, enabling the attention module to capture a comprehensive global receptive field while preserving precise positional information. This module helps the network accurately localize the target of interest. $z_c$ is the output of the c-channel, $H$ and $W$ represent the number of width and height of the input vector, respectively. $x_c(i,j)$ is the input vector of c-channel, with the i-th row and the j-th column.

$$z_c = \frac{1}{H \times W} \sum_{i=1}^{H} \sum_{j=1}^{W} x_c(i,j) \tag{1}$$

The process of generating coordinate attention utilizes the rich features generated through the embedding of coordinate information. During this step, the feature map derived from Eq 1 is further processed along both horizontal and vertical directions to produce intermediate feature maps $\mathbf{f} \in R^{C/r \times (H+W)}$ that encode detailed spatial information. $r$ is a reduction ratio to control the block's size. Then, the tensor $\mathbf{f}$ is decomposed into two different tensors $\mathbf{f}^h \in R^{C/r \times H}$ and $\mathbf{f}^w \in R^{C/r \times W}$. Subsequently, Eq 2 is used to process $\mathbf{f}^h$ and $\mathbf{f}^w$ independently, generating attention vectors $\mathbf{g}^h$ and $\mathbf{g}^w$.

$$\begin{aligned} \mathbf{g}^h &= \sigma\left(F_h\left(\mathbf{f}^h\right)\right) \\ \mathbf{g}^w &= \sigma\left(F_w\left(\mathbf{f}^w\right)\right) \end{aligned}, \tag{2}$$

where, $\sigma(\cdot)$ is a sigmoid activation function, and $F_h(\cdot)$ and $F_w(\cdot)$ represent 1 x 1 convolutional transformation function.

Finally, the output of the coordinate attention block can be represented by Eq 3, where $y_c(i,j)$ represents the output vector of c-channel with the i-th row and the j-th column, $x_c(i,j)$ represents the input vector of c-channel with the i-th row and the j-th column, $g_c^h(i)$ represents the attention vector in the x direction of c-channel in the i-th row and $g_c^w(j)$ represents the attention vector in the y direction of c-channel in column j:

$$y_c(i,j) = x_c(i,j) \times g_c^h(i) \times g_c^w(j) \tag{3}$$

*3.4. WIoUv3 loss function*

To mitigate the impact of low-quality samples on the model's training and generalization ability,

we introduce a WIoUv3 loss function in this study. WIoUv3 extends the WIoUv1 by incorporating a dynamic non-monotonic focusing mechanism (FM). First, let's provide an overview of WIoUv1 before delving into WIoUv3.

WIoUv1 is designed to alleviate the penalties associated with geometric factors like distance and aspect ratio, particularly for low-quality samples. This adjustment aims to reduce the influence of the loss function on the model during the training process. WIoUv1 utilizes a dual-attention mechanism, as shown in the following equations, $\mathcal{L}_{IoU}$ represents the loss in the intersection over union ratio between the predicted box and the true box:

$$\mathcal{L}_{WIoU\ v1} = R_{WIoU} \mathcal{L}_{IoU} \tag{4}$$

$$R_{WIoU} = \exp(\frac{(x-x_{gt})^2 + (y-y_{gt})^2}{(W_g^2 + H_g^2)^*}) \tag{5}$$

where $R_{WIoU} \in [1, e)$ is used to amplify the regular quality anchor boxes, $\mathcal{L}_{IoU} \in [1, e)$ is used to reduce the high-quality anchor boxes' $R_{WIoU}$, and when the predicted box has a good overlap with the ground truth box, the focus is placed on the distance to the center point.

As shown in Figure 5, $W_g$ and $H_g$ represent the size of the minimum enclosing box. $W_g$ and $H_g$ are detached from the computational graph (indicated by the superscript $*$) to prevent $R_{WIoU}$ generating gradients that hinder convergence.

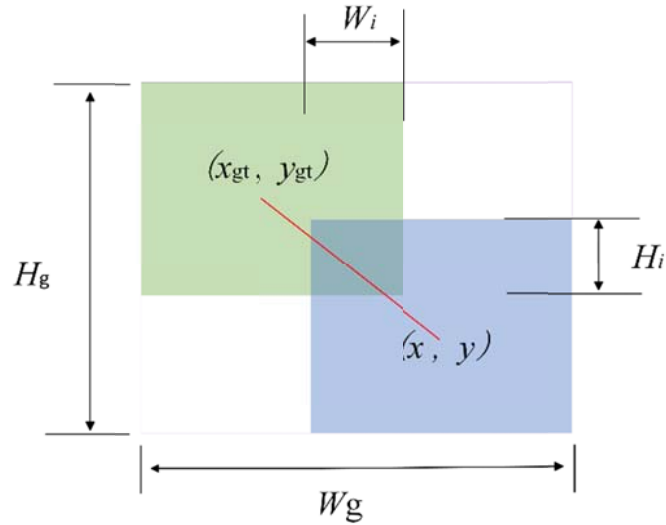

**Figure 5.** Schematic of the WIoU loss function.

WIoUv3 incorporates dynamic non-monotonic FM based on WIoUv1. Dynamic non-monotonic FM adopts a gradient gain allocation strategy and combines outliers and IoU loss for loss calculation.

Since $\mathcal{L}_{IoU}$ is dynamic, the quality division criterion for anchor boxes is also active, allowing WIoUv3 to adopt a gradient gain allocation strategy that best suits the current situation. The gradient gain allocation strategy reduces the competitiveness of high-quality anchor boxes while also reducing the harmful gradients generated by low-quality anchor boxes, allowing WIoUv3 to focus on regular-quality anchor boxes and improve the overall performance of the detector. The formula for WIoUv3 is as follows, $\delta$ and $\alpha$ are hyper-parameters that used to control outlier degree $\beta$ and gradient gains $r$:

$$\mathcal{L}_{WIoU\ v3} = r\mathcal{L}_{WIoU\ v1} \tag{6}$$

$$\text{where}\quad \beta = \frac{\mathcal{L}_{IoU}^*}{\mathcal{L}_{IoU}} \in [0, +\infty) \tag{7}$$

$$\text{where}\quad r = \frac{\beta}{\delta\alpha^{\beta-\delta}} \tag{8}$$

## 4. Experiment and analysis

Yunnan Power Grid Co., Ltd. provided the insulation defect dataset. It consists of 3000 sample images, with 1000 images for each of the three categories: normal, self-explosion missing, and partial damage. These 3000 images were randomly divided into training, validation, and test sets in a ratio of 6:2:2. The images are all taken by drones with a resolution of 4000 × 3000 per image. Since each image has multiple labels, the real label for each type is around 1600. The images in the dataset are all original images, without data augmentation and preprocessing.

*4.1. Experimental environment*

The experimental hardware and software utilized in this study comprises an i7-10700K CPU, a 32GB RAM and a NVIDIA RTX2080Ti GPU. The software environment was Windows 10, CUDA 10.2, PyTorch 1.10.0 and PyCharm Community Edition 2021.3.

*4.2. Experimental evaluation indicators*

The ablation experiments in this study utilized seven evaluation metrics: precision, recall, mAP_0.5, mAP_0.5:0.95, parameters, GFLOPS, and speed. The comparative experiments employed five evaluation metrics: precision, recall, mAP_0.5, mAP_0.5:0.95, and speed. The definitions of these metrics can be found in the reference [30].

*4.3. Experimental results and analysis*

4.3.1. Model training comparison

The experiment was carried out over 100 training epochs. Figures 6, 7, 8 and 9 show the comparison results of mAP.0.5, mAP0.5:0.95, precision, and recall during the training process, respectively. These visualizations clearly demonstrate the enhanced performance of our model when compared to the original model.

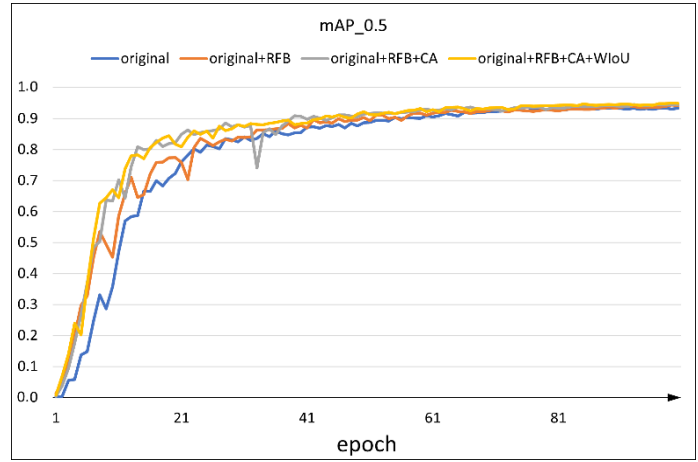

**Figure 6.** Comparison of mAP_0.5.

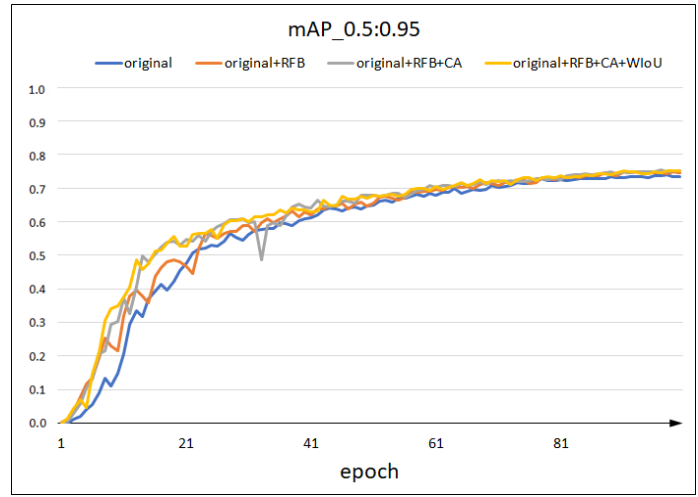

**Figure 7.** Comparison of mAP_0.5:0.95.

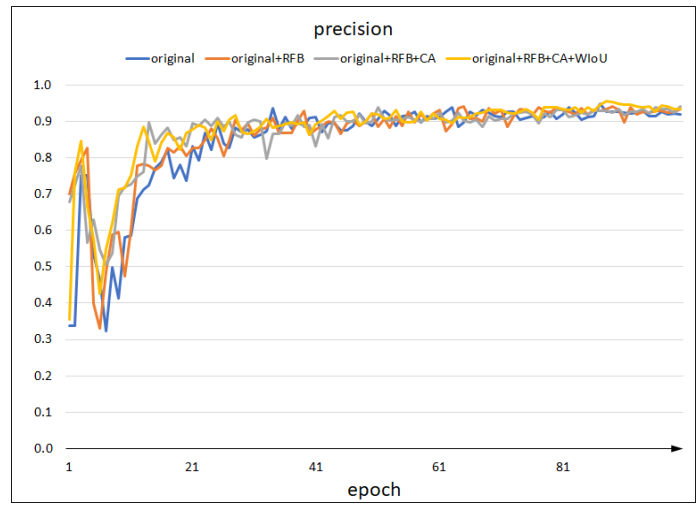

**Figure 8.** Comparison of precision.

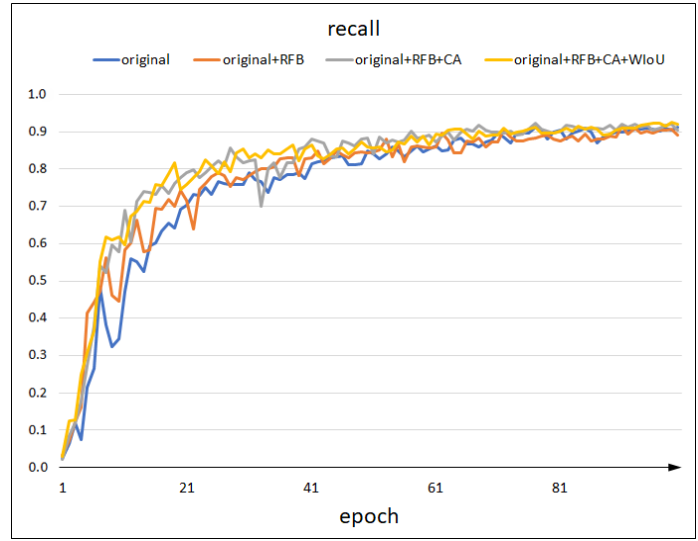

**Figure 9.** Comparison of recall.

4.3.2. Detection result

　　Figure 10 displays the detection results of the original YOLOv7 model. Subfigures (a), (b) and (c) correspond to partial damage detection, while subfigures (d), (e) and (f) correspond to self-explosion defect. On the other hand, each corresponding subfigures in Figure 11 display the detection results of our improved YOLOv7 model.

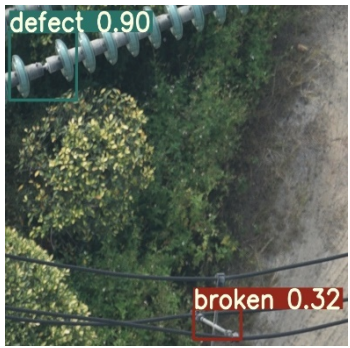 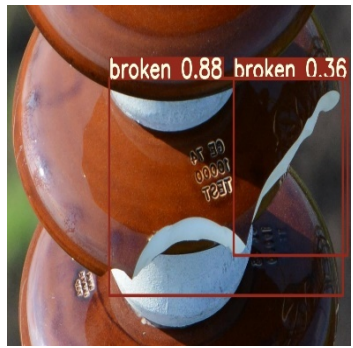 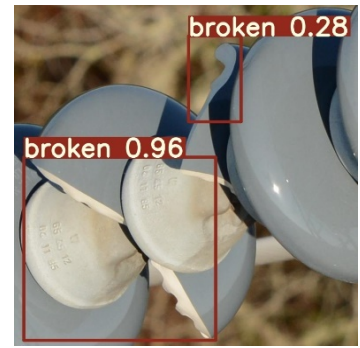

（a）Partial damage 1　　　　（b）Partial damage 2　　　　（c）Partial damage 3

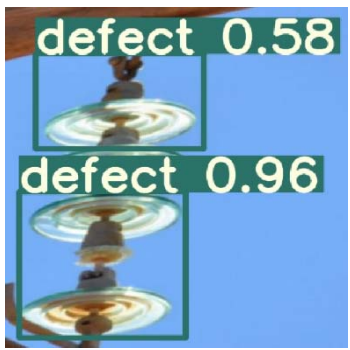 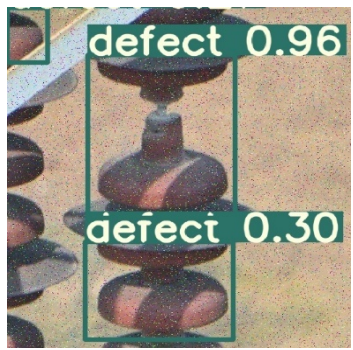 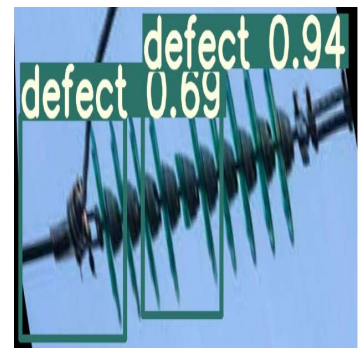

（d） Self-explosion defect 1（e）Self-explosion defect 2 （f）Self-explosion defect 3

**Figure 10.** Detection results of the original YOLOv7 model.

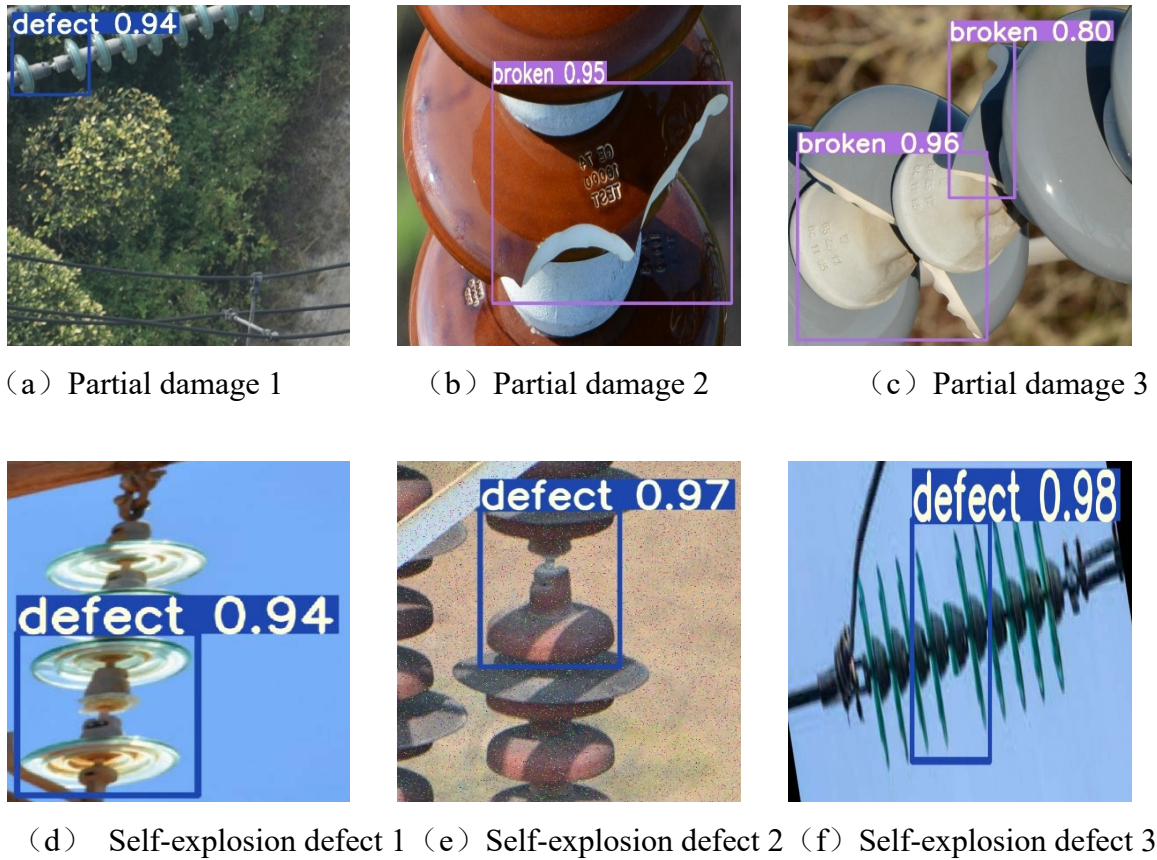

（a） Partial damage 1　　　　（b）Partial damage 2　　　　（c）Partial damage 3

（d） Self-explosion defect 1（e）Self-explosion defect 2（f）Self-explosion defect 3

**Figure 11.** Detection results of our improved YOLOv7 model.

Because the detected results have large sizes with a resolution of 4000 * 3000 pixels and the detected insulator is only a tiny proportion of the whole image, it isn't easy to see the actual label after detection. In Figure 11, we only intercepted the defective part of the insulator and magnified it for clear display in the paper.

Comparing the detection results of each subfigure in Figure 10 and Figure 11, our improved model does not produce false detections for partial damage and improves the confidence in detecting self-explosion loss targets in subfigure (a). In subfigure (b), our improved model does not have the same predictions for two parts of the same damaged insulator and exhibits higher detection confidence. In subfigure (c), our improved model shows more precise localization for partial damage. In subfigures (d), (e), and (f), our improved model does not produce false detections for self-explosion loss targets and improves confidence in detecting them.

4.3.3. Ablation experiments

The ablation experiments are performed to illustrate the efficacy of our enhancement. Table 1 presents the results.

**Table 1.** Results of the ablation experiments.

| RFB | CA | WIoU | Precision (%) | Recall (%) | mAP 0.5(%) | mAP0.5 :0.95(%) | Parameters | GFLOPs | Speed (ms) |
|---|---|---|---|---|---|---|---|---|---|
|  |  |  | 92.0 | 91.1 | 93.3 | 73.5 | 37.2M | 105.1 | 36.9 |
|  | √ |  | 91.6 | 90.5 | 93.8 | 73.4 | 37.7M | 105.2 | 38.4 |
|  |  | √ | 82.4 | 91.8 | 94.2 | 74.8 | 37.2M | 105.2 | 37.5 |
| √ |  |  | 93.4 | 89.1 | 94.1 | 74.6 | **33.9M** | **102.5** | **31.2** |
| √ | √ |  | **94.0** | 90.5 | 94.5 | 75.1 | **33.9M** | 102.6 | 32.2 |
| √ | √ | √ | 93.3 | **92.1** | **94.9** | **75.1** | 34.0M | 102.6 | 34.1 |

From Table 1, replacing the SPPCSPC module of the original YOLOv7 model with the RFB module improves mAP_0.5 by 0.8%, mAP_0.5:0.95 by 1.1%, precision by 1.4%, reduces parameters by 3.3M, reduces computation by 2.6 GFLOPS, and increases single-image detection speed by 5.7ms.

By incorporating both the RFB and CA attention mechanisms, mAP_0.5 improves by 1.2%, mAP_0.5:0.95 improves by 1.6%, precision improves by 2%, parameters reduce by 3.3M, computation reduces by 2.5 GFLOPS, and single-image detection speed increases by 4.7ms.

Furthermore, by including the RFB, CA attention mechanism, and WIoU loss function simultaneously, mAP_0.5 improves by 1.6%, mAP_0.5:0.95 improves by 1.6%, precision improves by 1.3%, recall improves by 1%, parameters reduce by 3.2M, computation reduces by 2.5 GFLOPS, and single-image detection speed increases by 2.8ms.

4.3.4. Attention visual comparison

To demonstrate that coordinate attention can better address the issue of small-scale insulator self-bursting and damage, we conducted a visual comparison between the improved model's CA and the attention of the original model. This comparison is shown in Figure 12.

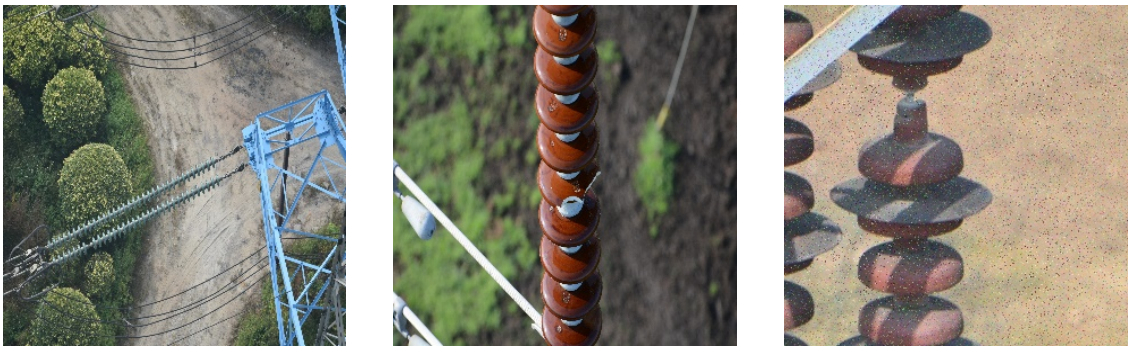

（a）Detected original samples

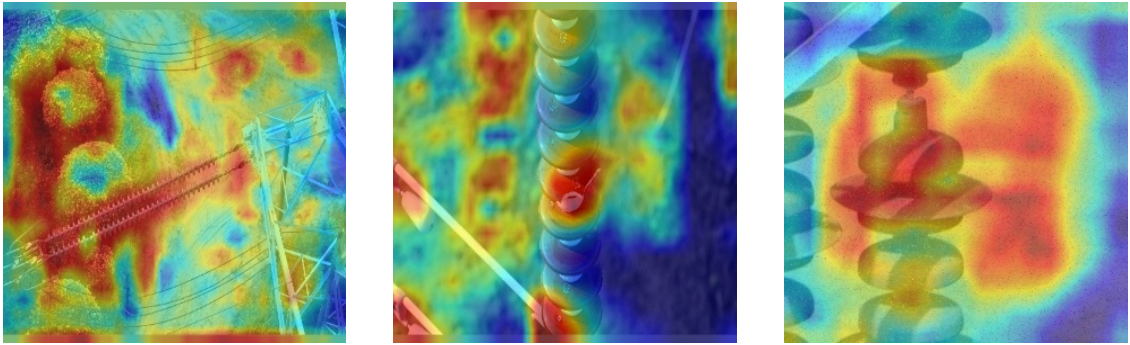

（b）Attention visualization using the original model

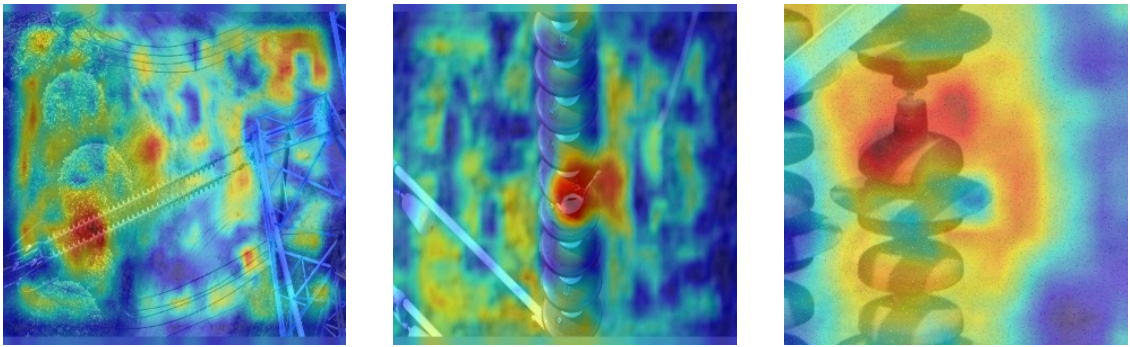

（c）Attention visualization using our improved model

**Figure 12.** Comparison of attentional visualization.

Based on the comparison between Figure 12(b) and Figure 12(c), it can be observed that coordinate attention can better focus on the small-scale regions of insulator self-bursting and damage.

4.3.5. Comparative experiments with other attention mechanisms

To verify the effectiveness of our attention mechanism, we added other attention mechanisms to the same branch of the original model for comparison.

**Table 2.** Comparative experiments with other attention mechanisms.

| Attention Mechanisms | Precision (%) | Recall (%) | mAP 0.5 (%) | mAP 0.5:0.95 (%) | Parameters | GFLOPs | Speed (ms) |
| --- | --- | --- | --- | --- | --- | --- | --- |
| Original | 92.0 | **91.1** | 93.3 | **73.5** | **37.2M** | **105.1** | **36.9** |
| CBAM | **92.4** | 87.6 | 92.1 | 69.2 | 42.0M | **105.1** | 58.1 |
| ECA | 90.5 | 86.2 | 90.9 | 68.3 | 37.5M | 105.3 | **38.6** |
| GAM | 89.0 | 88.3 | 91.6 | 69.4 | 53.8M | 111.6 | 62.6 |
| SimAM | 90.6 | **91.1** | 93.4 | 72.9 | **37.2M** | **105.1** | 39.5 |
| CA | 91.6 | 90.5 | **93.8** | 73.4 | 37.7M | 105.2 | 38.6 |

As shown in Table 2, by adding various attention mechanism at the same location of the model, the CA attention mechanism achieves the highest accuracy on mAP0.5 and mAP0.5:0.95 is higher than other attention mechanisms.

4.3.6. Comparative experiments with other IoU loss functions

We conduct comparative experiments with other loss functions to verify the effectiveness of the WIoUv3 loss function. Total loss is the sum of box_loss, obj_class, and cls_loss.

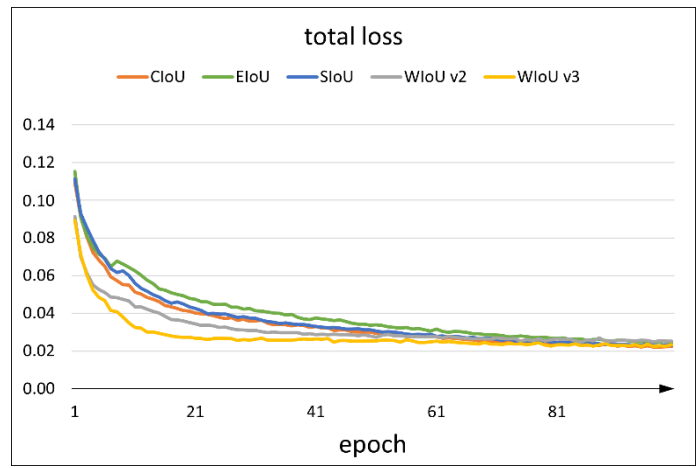

**Figure 13.** Comparison of IoU loss function.

As shown in Figure 13, we use the WIoUv3 loss function to make our model converge faster and reach stability than the CIoU loss function used by the YOLOv7 model.

4.3.7. Experimental comparison with different models

The comparison experiment with the mainstream model has been implemented, which has verified the progressiveness of our improved model. YOLOv8, YOLOv6, and YOLOv5 have multiple versions with varying network depths and widths. Due to constraints in our GPU resources, we selected a subset of versions (m versions) of YOLOv8, YOLOv6, and YOLOv5 for comparison. The performance of other versions can be estimated proportionally. The outcomes of the comparative experiments are presented in Table 3.

**Table 3.** Experimental comparison with different models.

| Model | Precision (%) | Recall (%) | mAP 0.5 (%) | mAP 0.5:0.95 (%) | Parameters | GFLOPs | Speed (ms) |
|---|---|---|---|---|---|---|---|
| Faster RCNN | 81.7 | 58.8 | 81.6 | 53.2 | 41.4M | 81.9 | 91.7 |
| Sparse RCNN | 78.2 | 63.8 | 76.5 | 51.2 | 106.1M | 64.6 | 98.9 |
| YOLOv5m | 93.1 | 90.2 | 93.0 | 70.6 | **21.2M** | 49.0 | 38.3 |
| YOLOv6m | 70.8 | 59.0 | 70.8 | 46.9 | 34.9M | 85.8 | 66.3 |

| | | | | | | |
|---|---|---|---|---|---|---|
| YOLOv7 | 92.0 | 91.1 | 93.3 | 73.5 | 37.2M | 105.1 | 36.9 |
| YOLOv8m | **93.6** | 88.6 | 93.6 | **76.8** | 25.9M | 78.9 | 61.7 |
| Ours | 93.3 | **92.1** | **94.9** | 75.1 | **34.0M** | 102.6 | **34.1** |

From Table 3, our model achieves the highest mAP_0.5 score, reaching 94.9%. Although it is not the highest in terms of mAP_0.5:0.95 and precision, it ranks second and achieves the most elevated values regarding recall and speed, reaching 92.1% and 34.06ms, respectively. Compared to other detection models, our model outperforms in terms of mAP_0.5, recall, and speed, achieving the highest performance.

4.3.8. Detection results for various defect types

Table 4 compares defect detection results between the Yolov7 and our improved model for each category.

**Table 4.** Defect detection results for various types.

| Model | Type | Labels | Precision (%) | Recall (%) | mAP 0.5 (%) | mAP 0.5:0.95 (%) |
|---|---|---|---|---|---|---|
| YOLOv7 | All | 956 | 92.0 | 91.1 | 93.3 | 73.5 |
| | Self-explosion | 302 | 94.0 | 83.4 | 89.1 | 60.6 |
| | Partial damage | 317 | 97.6 | 93.5 | 95.1 | 78.6 |
| | Normal | 337 | 84.4 | 96.4 | 95.8 | 81.5 |
| Ours | All | 956 | 93.3 | 92.1 | 94.9 | 75.1 |
| | Self-explosion | 302 | 96.0 | 85.6 | 91.5 | 63.5 |
| | Partial damage | 317 | 97.8 | 93.6 | 96.7 | 79.4 |
| | Normal | 337 | 86.2 | 97.2 | 96.6 | 82.5 |

Table 4 illustrates that our enhanced model surpasses the original YOLOv7 model across all three detection types in terms of precision, recall, mAP0.5 and mAP0.5:0.95.

## 5. Conclusion

In this study, we propose an enhanced model based on YOLOv7 for detecting various types of insulator defects. Our model can automatically detect aerial images of normal, self-exploded missing, and partially damaged insulators, reducing the workload of insulator inspection and improving inspection efficiency.

Three difficulties exist in insulator defect detection: the diversity of insulator materials and colors, complex backgrounds, and diverse damages. This study uses the RFB module to enhance the network's feature extraction capability, incorporates the CA mechanism to improve small target detection, and introduces the WIoU loss function to address low-quality samples hindering model

44